\documentclass[letterpaper,10pt,comsoc]{ieeeconf}  % Comment this line out if you need a4paper
\usepackage[T1]{fontenc}
\usepackage{todonotes}

\usepackage{amsmath, amssymb}
\usepackage{mathtools}
\usepackage{physics}

\usepackage{array}
\newcommand{\PreserveBackslash}[1]{\let\temp=\\#1\let\\=\temp}
\newcolumntype{C}[1]{>{\PreserveBackslash\centering}p{#1}}
\newcolumntype{R}[1]{>{\PreserveBackslash\raggedleft}p{#1}}
\newcolumntype{L}[1]{>{\PreserveBackslash\raggedright}p{#1}}

\usepackage{siunitx}
\usepackage{subcaption,graphicx}

\usepackage[normalem]{ulem}
\usepackage{hyperref}

\usepackage{cleveref,tabularx}
\Crefname{figure}{Fig.}{Figs.}
\Crefname{section}{Sec.}{Secs.}
\Crefname{table}{Tab.}{Tabs.}
\Crefname{equation}{}{}

\newcommand{\df}{dorsiflexion}

\definecolor{forestgreen}{RGB}{34 139 34}
\definecolor{darkorchid}{RGB}{153 50 204}
\definecolor{darkorange}{RGB}{255 140 0}

\graphicspath{{./image/}}

\title{\LARGE \bf 
Gastrocnemius and Power Amplifier Soleus Spring-Tendons\\Achieve Fast Human-like Walking in a Bipedal Robot
}
\author{Bernadett Kiss$^{1}$, Emre Cemal Gonen$^{1}$, An Mo$^{1}$, Alexandra Buchmann$^{2}$,\\ Daniel Renjewski$^{2}$ and Alexander Badri-Spr\"owitz$^{1}$% <-this % stops a space
\thanks{$^{1}$Dynamic Locomotion Group, Max Planck Institute for Intelligent Systems, 70569 Stuttgart, Germany \\ {\tt \footnotesize [kiss;gonen;mo;sprowitz]@is.mpg.de}}%
\thanks{$^{2}$Chair of Applied Mechanics, TUM School of Engineering \& Design, Department of Mechanical Engineering, Technical University of Munich, 85748 Garching near Munich, Germany {\tt \footnotesize [firstname.lastname]@tum.de}}%
}
\begin{document}
\maketitle
\thispagestyle{empty}
\pagestyle{empty} % empty % change for submission back to empty, plain to show the page numbers at the center
%
%------------------------------------------------------------------------------
%ABSTRACT:
%------------------------------------------------------------------------------
\begin{abstract}
% Brief summary of your paper, about one sentence per section.
Legged locomotion in humans is governed by natural dynamics of the human body and neural control. One mechanism that is assumed to contribute to the high efficiency of human walking is the impulsive ankle push-off, which potentially powers the swing leg catapult. However, the mechanics of the human lower leg with its complex muscle-tendon units spanning over single and multiple joints is not yet understood. Legged robots allow testing the interaction between complex leg mechanics, control, and environment in real-world walking gait. We developed a \SI{0.49}{m} tall, \SI{2.2}{kg} anthropomorphic bipedal robot with Soleus and Gastrocnemius muscle-tendon units represented by linear springs, acting as mono- and biarticular elastic structures around the robot's ankle and knee joints. We tested the influence of three Soleus and Gastrocnemius spring-tendon configurations on the ankle power curves, the coordination of the ankle and knee joint movements, the total cost of transport, and walking speed. We controlled the robot with a feed-forward central pattern generator, leading to walking speeds between \SI{0.35}{m/s} and \SI{0.57}{m/s} at \SI{1.0}{Hz} locomotion frequency, at \SI{0.35}{m} leg length. We found differences between all three configurations; the Soleus spring-tendon modulates the robot's speed and energy efficiency likely by ankle power amplification, while the Gastrocnemius spring-tendon changes the movement coordination between ankle and knee joints during push-off.

% \vspace*{5cm}
\end{abstract}
%Use this command to check your structure:
%\tableofcontents
%------------------------------------------------------------------------------
%INTRODUCTION:
%------------------------------------------------------------------------------
%
\section{INTRODUCTION}
%\begin{itemize}
%\item What is the context / motivation of your work?
%\item What is the State of the Art? Cite relevant literature.  \cite{IEEEexample:shellCTANpage}
%EXAMPLE FOR CITATIONS: 
%Example for a citation: The principle of complementary limb motion estimation assumes that an individual's movement intention for missing or paralyzed limbs can be estimated from residual body motion~\cite{ValTNSRE2008}.
%remark: "~" prevents a line break
%\item What is missing in the literature? 
%\item What is your new contribution?
%\end{itemize}
The complexity of human bipedal walking becomes apparent when attempting to technically replicate or restore the human lower limb's musculoskeletal system, e.g., with humanoid robots or lower limb prostheses. Robots and prostheses do not yet reach human performance in terms of efficiency, mobility, and robustness. The current gap implies a missing understanding of biomechanics and control of human locomotion. 

One mechanism assumed to contribute to the high efficiency of human walking is the impulsive ankle push-off, which potentially powers the human swing leg catapult ~\cite{Lipfert2014}. A catapult's function is physically characterized by slow storage of elastic energy, followed by a rapid release of stored energy, with a substantially higher output power that accelerates a projectile. Hof et al.\, described first how the ankle power burst in the late stance phase is preceded by a slower energy storage phase in human walking \cite{Hof1983}. The observed ankle power burst is hereby higher than the plantar flexor muscles' peak power \cite{Lipfert2014,Fukunaga.2001,Ishikawa.2005}. Thus, the excessive power burst indicates that additional passive structures store elastic energy which is transformed into kinetic energy during push-off \cite{Hof1983,Alexander.1991}.  
The rapid release of the stored elastic energy at the end of stance is available to accelerate the stance leg into swing like a catapult firing its projectile \cite{Lipfert2014}. 

A catapult has three main mechanical components: an elastic element, a block, and a catch with or without escapement. In the human lower leg, the complex interplay between the thigh-shank-foot segment chain and muscle-tendon units (MTUs) spanning the ankle joint makes it hard to identify the exact swing leg catapult mechanics and its functional elements. Elastic energy storage in stance and rapid recoil during push-off is facilitated by the Achilles tendon attached to the Soleus (SOL) and Gastrocnemius (GAS) muscles \cite{Fukunaga.2001,Ishikawa.2005,Lichtwark.2006,Lichtwark.2007}. The GAS muscle works mostly isometrically during stance, facilitating energy efficient loading of the Achilles tendon. In contrast, the SOL muscle seems to contribute to ankle coordination by active contraction \cite{Cronin.2013c}. Up to \SI{91}{\%} of the power output during push-off is provided by elastic energy \cite{Cronin.2013}. Changes in the effective ankle stiffness influence the energy efficiency of walking, as studies with passive ankle foot orthoses showed \cite{Sawicki.2009,Collins.2015}. The combination of ground and foot is the human leg catapult's block, against which the swing leg catapult unloads at the end of single support, after shifting the body weight from the stance to the leading leg. Unloading happens just before the leading leg touches the ground, i.e., in free fall. The catch of the human swing leg catapult has been hypothesised but not functionally described yet.

As part of the elastic element of the catapult, the biarticular GAS MTU could have a different function from the monoarticular SOL MTU. Advantages of multi-articular actuators and spring-tendons have been shown in robots. The BioBiped3 robot with biarticular muscles showed improved balance control during upright standing and locomotion, and axial leg function during bouncing \cite{sharbafi_new_2016}. CARL robot improved its hopping efficiency by \SI{16}{\%} by using multi-articular actuators~\cite{nejadfard_design_2019}. A simulation study showed that combining mono- and bi-articular foot prosthesis actuation reduces peak power requirements \cite{eslamy_adding_2014}. BirdBot's spring-tendon network improved cost of transport, and enabled robust engagement and disengagement of the leg's parallel elasticity~\cite{birdbot_2022}. So far, the exact role and function of plantarflexor spring-tendons as part of the swing leg catapult during walking has not been studied in robots. 

\begin{figure*}[!h]
\centering
\includegraphics[width=\linewidth]{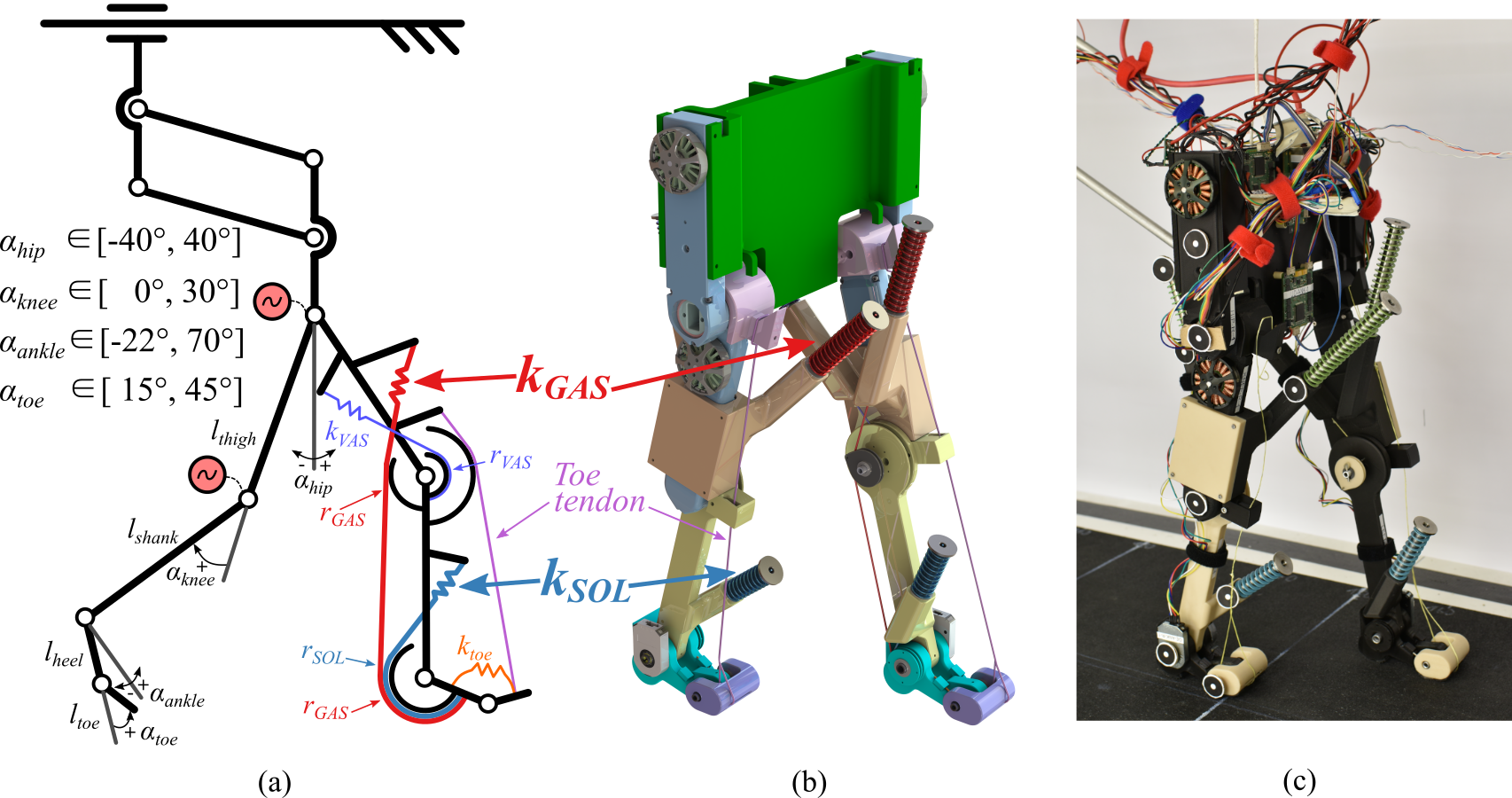}
\caption{
a) Schematic of the robot with spring-tendon routing, angle definitions, and range of motion of the joints. $l_{thigh}=l_{shank}=\SI{160}{mm}$, $l_{heel}=\SI{48}{mm}$, $l_{toe}=\SI{24}{mm}$, $r_{GAS}=r_{SOL}=\SI{13}{mm}$, $r_{VAS}=\SI{12}{mm}$, $k_{VAS}=\SI{2.16}{N/mm}$, $k_{toe}=\SI{8.04}{Nmm/deg}$. For further segment length and weight ratios see \Cref{tab:segmentdata}. The toe tendon (pink) is not connected to a spring but is mounted slack during stance. It features a stiffness of $\SI{60}{N/mm}$.
b) Rendering of the bipedal robot.
c) Picture of the bipedal robot. The robot's trunk movement is constrained to translations in the sagittal plane, and all trunk rotations are locked. The four-bar guide slides freely in fore-aft direction.
}
\label{fig:model}
\end{figure*}
In this work we show the influence of SOL and GAS muscles on a anthropomorphic bipedal robot's ankle power curve, ankle and knee joint movement coordination, energy efficiency and walking speed. We analyze the robot's energy consumption for one gait cycle and report the observed ankle power amplification of the swing leg catapult, inspired by human locomotion research like \cite{Lipfert2014,Fukunaga.2001,Ishikawa.2005,Lichtwark.2006}. As testing platform, the developed robot features both ankle muscles as linear springs.

Developing a deeper understanding of the swing leg catapult and its functional components will help to increase the efficiency of walking for legged robots and bears great potential to improve gait rehabilitation devices and prosthesis toward a more natural gait pattern. 
\begin{table}[t]
\caption{Robot ($GAS+SOL$) and human segment length and weight ratio comparison.}
\label{tab:segmentdata}
\begin{center}
\begin{tabular}{llccc}
 Length ratios & & Robot & Human \cite{Plagenhoef1983} & Human \cite{Gharooni2007}	\\
\hline
&Trunk       & 5.09 & 6.94 & 7.38 \\
&Thigh       & 5.82 & 5.66 & 6.28 \\
&Shank       & 5.82 & 5.93 & 6.31 \\
&Foot length & 3.49 & N/A  & 3.90 \\
&Foot height & 1.00 & 1.00 & 1.00 \\
%\hline
Weight ratios & & & & \\
\hline
&Trunk & 9.42 & 39.24 & 34.37  \\
&Thigh & 6.68 & 8.06  & 6.91  \\
&Shank & 1.76 & 3.66  & 3.21  \\
&Foot  & 1.00 & 1.00  & 1.00   \\
\end{tabular}
\end{center}
\end{table}
%
%------------------------------------------------------------------------------
%METHODS:
%------------------------------------------------------------------------------
\section{METHODS}
%The methods section contains the recipe that allows others to reproduce your results, without reading the results section. 
%So, here you answer questions like:
%\begin{itemize}
%\item How do you design your device/your controller? % The latter could e.g. be put in a \subsection{Control Design}
%\item How do you design experiments/simulations? % E.g. \subsection{Experimental Protocol}
%\item How do you collect and analyze data? % E.g. \subsection{Data Acquisition}, \subsection{Data Analysis}
%\end{itemize}
%Choose descriptive names for new methods (instead of words like ``novel''), to simplify citation and future improvements.
%
%
We designed an anthropomorphic bipedal robot, the size of a small child. The robot's hip and knee joints were actuated and controlled by open-loop central pattern generators \cite{ijspeert_central_2008}. The control patterns follow anthropomorphic joint trajectories. The robot's ankle joints were not actuated. To gain insights into the separate roles of SOL and GAS MTUs acting at the ankle and knee joints, SOL and GAS MTUs were represented by linear springs connected to tendons acting on pulleys of fixed radii around the joints (\Cref{fig:model}). We tested three spring-tendon configurations: $GAS+SOL$, $SOL$ and $GAS$. 
\begin{table}[b]
\caption{SOL and GAS spring stiffness, and robot mass values in the three robot configurations.}
\label{tab:robot_para}
\begin{center}
\begin{tabular}{L{2.7cm}C{1.5cm}C{1.3cm}C{1.3cm}}
 						& GAS + SOL     & SOL	    &  GAS \\
\hline
SOL Stiffness   [N/mm]   & 4.5          & 6.1      & 0     \\
GAS Stiffness   [N/mm]   & 1.4           & 0         & 6.1  \\
Robot mass      [kg]    & 2.22         & 2.05     & 2.05 \\
\end{tabular}
\end{center}
\end{table}
\subsection{Bipedal robot design}
Our humanoid robot features passive toe and ankle joints, actuated knee joints with parallel elasticity, and actuated hip joints (\Cref{fig:model}). The robot's actuated segments for hip and knee were adapted from the SOLO robot module which is described in detail in \cite{Grimminger2019}. Segment length ratios (\Cref{tab:segmentdata}) and joint range of motions (\Cref{fig:model}) of the robot were designed according to human data \cite{Plagenhoef1983, Gharooni2007} and 2D neuromuscular simulation studies of human walking \cite{Geyer.2010}. 

The robot's elasticities represent muscle-tendon units in the human leg and foot. Spring stiffness values were based on initial approximations which are discussed subsequently (\Cref{tab:robot_para}, \Cref{fig:model}).

% SOL and GAS
SOL and GAS spring-tendons spanning the ankle joint are loaded from the maximum ankle plantarflexion angle of \SI{22}{\degree} until the maximum ankle dorsiflexion angle of \SI{15}{\degree}. At peak \df{} angle both ankle spring-tendons together produce a torque that supports one third of the robot's total weight. 

% VAS; 
The VAS spring, representing the vasti muscle group of the quadriceps, spans the knee joint acting on its knee pulley radius $r_\mathrm{VAS}$. The VAS spring induces knee torque in parallel to the knee motor, similar to \cite{ashtiani_hybrid_2021}. The VAS spring is loaded starting from \SI{0}{\degree} knee joint angle, and produces a torque that supports \SI{10}{\%} of the robot's total weight at \SI{20}{\degree} knee flexion angle.

% TOE SPRINGS
Rotational toe springs are loaded in stance from an initial toe angle of \SI{15}{\degree}. At the maximum toe angle of \SI{45}{\degree}, the toe spring produces a torque that supports one third of the robot's total weight.

% TOE TENDON
Additionally, a toe tendon (without spring) was attached between the thigh and toe segments which pulled up the toes, i.e., facilitated toe clearance in swing when the knee was flexed more than \SI{20}{\degree} (\Cref{fig:model}). The toe tendon pulled up the toes only in swing since the knee is not flexed more than \SI{20}{\degree} in stance phase \cite{Perry.2010}.
\subsection{Robot control}
% important characteristics of solo module ?
% CPG control based on neuromuscular simulation results on joint angles
% The reference trajectories for hip and knee we used to design the CPG control were extracted from a 2D neuromuscular simulation of human walking \cite{Geyer.2010}. 
% Hybrid CPG control (ref)
% Actuator modules consist of brushless DC motors adapted from the open-source quadruped robot SOLO \cite{Grimminger2019}. Each module features a 9:1 belt transmission, a high-resolution optical encoder, and a lightweight 3D-printed shell that forms a leg segment. With its low gear transmission ratio, we were able to reach high actuator velocities, and accurate torque control using SOLO's motor controller in current control mode.
%
\begin{table}[b]
\caption{CPG control parameters.}
\label{tab:cpg_para}
\begin{center}
\begin{tabular}{llccc}
Parameter 					& 		 	         &  &  Value  &	 \\
        					& 		 	         & $GAS+SOL$ & $SOL$   & $GAS$	 \\
        					
\hline
Frequency                   & $f$		            &  \SI{1}{Hz}  & \SI{1}{Hz} &  \SI{1}{Hz} \\
Hip duty factor 			& $D_\text{hip}$		&  0.65  & 0.65 & 0.65 	\\
Knee duty factor 			& $D_\text{knee}$		& 0.65 & 0.65 & 0.60 	\\
Hip amplitude 				& $\Theta_\text{hipAmplitude}$ 		& \SI{30}{\degree}  &\SI{30}{\degree} & \SI{27.5}{\degree} 	\\
Knee amplitude 				& $\Theta_\text{kneeAmplitude}$		& \SI{70}{\degree} & \SI{70}{\degree} & \SI{70}{\degree} \\
Hip offset 					& $\Theta_\text{hipOffset}$			& \SI{5}{\degree} &  \SI{2.5}{\degree} & \SI{2.5}{\degree} 	\\
Knee offset 				& $\Theta_\text{kneeOffset}$		& \SI{2}{\degree}  & \SI{2}{\degree} & \SI{2}{\degree} 	\\
Hip swing steady            & $\varphi_\text{hip}$			                    & 0.05  & 0.05 & 0.05 	\\
\end{tabular}
\end{center}
\end{table}

We implemented Central Pattern Generators (CPGs) for each actuated joint that control the bipedal walking gait. The CPGs are modified Hopf oscillators modeled in phase space \cite{ijspeert2007swimming}. The walking gait was implemented with a half-cycle phase shift between left and right oscillator nodes. 

CPG joint trajectories were controlled with eight parameters. The hip offset ($\Theta_\text{hipOffset}$) and amplitude ($\Theta_\text{hipAmplitude}$) describe the hip trajectory. The knee offset ($\Theta_\text{kneeOffset}$) and amplitude ($\Theta_\text{kneeAmplitude}$) define the knee trajectory. The frequency ($f$) describes the robot's overall speed, and duty factors ($D_\text{hip}$ and $D_\text{knee}$) define the ratio between stance duration and swing duration for hip and knee separately. The hip swing steady parameter ($\varphi_\text{hip}$) defines the hip's steady duration at the end of the swing, as a fraction of a gait cycle.

The CPG is described by a set of coupled differential equations:
\begin{equation}
\label{eq:oPhasedot_definition}
    \dot{\mathbf{\Phi}}_i = \mathbf{\Omega}_i
\end{equation}
where $\mathbf{\Phi}_i$ is the oscillator phase vector for leg $i$, and $\mathbf{\Omega}_i$ is the angular velocity vector for leg $i$. Since the bipedal robot has two legs, there were two oscillator nodes. These nodes generate oscillator phase values to be used to determine the joint angle values for each leg. Furthermore, the phase shift between these two nodes is desired to be $\pi$ to ensure one leg swings forward while the other leg swings backward. The correlation between these two nodes ($N = 2$) is defined using a coupled relation based on the desired phase shift, as shown in the following equation:
\begin{equation}
\label{eq:oPhasedot_formula}
    \dot{\mathbf{\Phi}}_i = 2 \pi f + \sum_{j = 1}^{N}{\alpha_\text{dyn,ij} \cdot \mathbf{C}_\text{ij} \cdot \sin{ \left( \mathbf{\Phi_\text{j}} - \mathbf{\Phi_\text{i}} - \mathbf{\Phi^\text{d}_\text{ij}} \right) }}
\end{equation}
where $f$ is the frequency, $\alpha_\text{dyn,ij}$ is the conversion constant of the network dynamics between nodes $i$ and $j$ (we use $\alpha_\text{dyn,ij} = 5$), $\mathbf{C}_\text{ij}$ is the coupling matrix weight between nodes $i$ and $j$, $\mathbf{\Phi^\text{d}_\text{ij}}$ is the desired phase difference matrix value between nodes $i$ and $j$.

The hip trajectory for a leg $i$ is a cosine wave, as in \Cref{eq:hip_trajectory}.
% The phase of the signal is varying and the function of the oscillator phase in node $i$. 
\begin{equation}
\label{eq:hip_trajectory}
    \Theta_\text{hip,i} = \Theta_\text{hipAmplitude,i} \cos{\phi_{i}} + \Theta_\text{hipOffset,i}
\end{equation}
where $\Theta_\text{hip,i}$ is the hip trajectory for the $i^\text{th}$ leg, $\Theta_\text{hipAmplitude,i}$ is the hip amplitude and $\Theta_\text{hipOffset,i}$ is the hip offset. The phase $\phi_{i}$ that is a function of the oscillator phase is defined as a continuous piece-wise function having three subdomains presented in \Cref{eq:end_effector_phase_hip}. These subdomains are defined as intervals based on the duty factor of the hip ($D_\text{hip}$) and the hip swing steady parameter ($\varphi_\text{hip}$) that are defining percentages of a gait cycle. Based on where the oscillator phase lies in the domain, $\phi_{i}$ can be determined accordingly.
\begin{equation}
\label{eq:end_effector_phase_hip}
    \phi_{i} = \begin{dcases} 
\frac{\mathbf{\Phi_\text{i}}}{2 D_\text{hip}} & \mathbf{\Phi_\text{i}} < 2 \pi D_\text{hip} \\ & \\
\frac{\mathbf{\Phi_\text{i}} + 2 \pi (1 - 2 D_\text{hip} - \varphi_\text{hip})}{2 (1 - D_\text{hip} - \varphi_\text{hip})} & 2 \pi D_\text{hip} \leq \mathbf{\Phi_\text{i}} \text{ and} \\
& \mathbf{\Phi_\text{i}} < 2 \pi (1 - \varphi_\text{hip})\\ & \\
2 \pi & \text{else}
 \end{dcases} 
\end{equation}
Similarly to hips, the desired knee trajectory is a sine-wave.
\begin{equation}
\label{eq:knee_trajectory}
    \Theta_\text{knee,i} = \Theta_\text{kneeAmplitude,i} \sin{\phi_{i}} + \Theta_\text{kneeOffset,i}
\end{equation}
The phase for the knee that is shaped by $D_\text{knee}$ was calculated using:
\begin{equation}
\label{eq:end_effector_phase_knee}
    \phi_{i} = \begin{dcases} 
0 & \mathbf{\Phi_\text{i}} < 2 \pi D_\text{knee} \\
\frac{\mathbf{\Phi_\text{i}} - 2 \pi D_\text{knee}}{2 (1 - D_\text{knee})} & \text{else}
 \end{dcases} 
\end{equation}
The reference trajectories for hip and knee, used to design the CPG control (\Cref{eq:hip_trajectory,eq:knee_trajectory}), were extracted from a 2D neuromuscular simulation of human walking \cite{Geyer.2010}. We initially set the CPG parameters based on this reference, and hand-tuned the parameters during the experiments toward a stable walking pattern (\Cref{tab:cpg_para}). The CPG output forms a position reference for PD controllers to calculate desired joint motor currents in the actuator modules \cite{Grimminger2019}:
\begin{equation}
\label{eq:control}
    \begin{split}
        i_\text{hipMotor} & = k_{p,\text{hip}} \Delta \Theta_\text{hip} + k_{d,\text{hip}} \dv{t} \Delta \Theta_\text{hip} \\
        i_\text{kneeMotor} & = k_{p,\text{knee}} \Delta \Theta_\text{knee} + k_{d,\text{knee}} \dv{t} \Delta \Theta_\text{knee}
    \end{split}
\end{equation}
where $i_\text{hipMotor}$ and $i_\text{kneeMotor}$ are the desired motor current values for hip and knee, respectively. $k_{p}$ and $k_{d}$ are the PD controller gains. For all of experiments, $k_{p,\text{hip}} = 30$ and $k_{d,\text{hip}} = 0.2$ are used for hip modules while $k_{p,\text{knee}} = 15$ and $k_{d,\text{knee}} = 0.2$ for knee modules. $\Delta \Theta$ is the error between the desired angle value and the current angle.

\begin{figure*}[!ht]
\centering
\includegraphics[width=\linewidth]{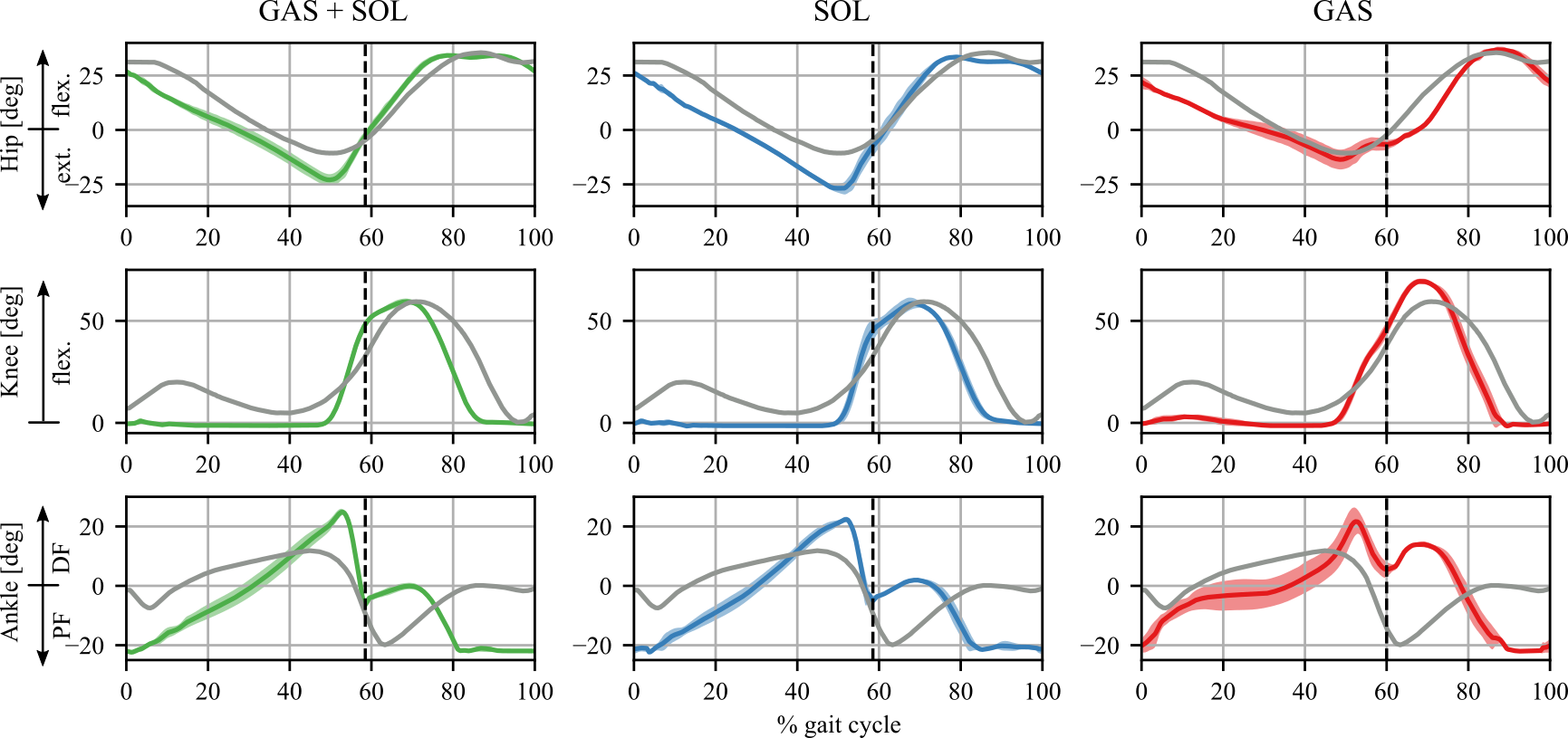}
\caption{Hip, knee and ankle joint angle curves during the three experimental configurations (in color) and human reference data (in gray, from \cite{Perry.2010}). Shading shows the standard deviation of the curves. Positive direction corresponds to flexion for hip and knee angles, and to dorsiflexion (DF) for the ankle angle. The vertical dashed lines indicate toe-off. The joint angles show no substantial changes between $GAS+SOL$ and $SOL$ configurations, while all three joint curves change in the $GAS$ configuration.}
\label{fig:angle}
\vspace{-0.45cm}
\end{figure*}

\subsection{Experimental setup}
We tested three robot ankle configurations: $GAS+SOL$, $SOL$, and $GAS$. 
In the first experimental configuration, the ratio between the ankle joint torques produced by the GAS and SOL spring-tendons was set to the physiological ratio between GAS and SOL MTU maximum isometric force, $1:3$ \cite{Delp1990} ($GAS+SOL$). In the second ($SOL$) and third ($GAS$) configurations, only SOL and only GAS spring-tendons were used to test the influence of SOL and GAS separately. We slightly adapted CPG parameters for each configuration to ensure stable robot gaits (\Cref{tab:cpg_para}).

\begin{figure*}[ht]
\centering
\begin{subfigure}[b]{.30\textwidth} \includegraphics[width=.9\textwidth]{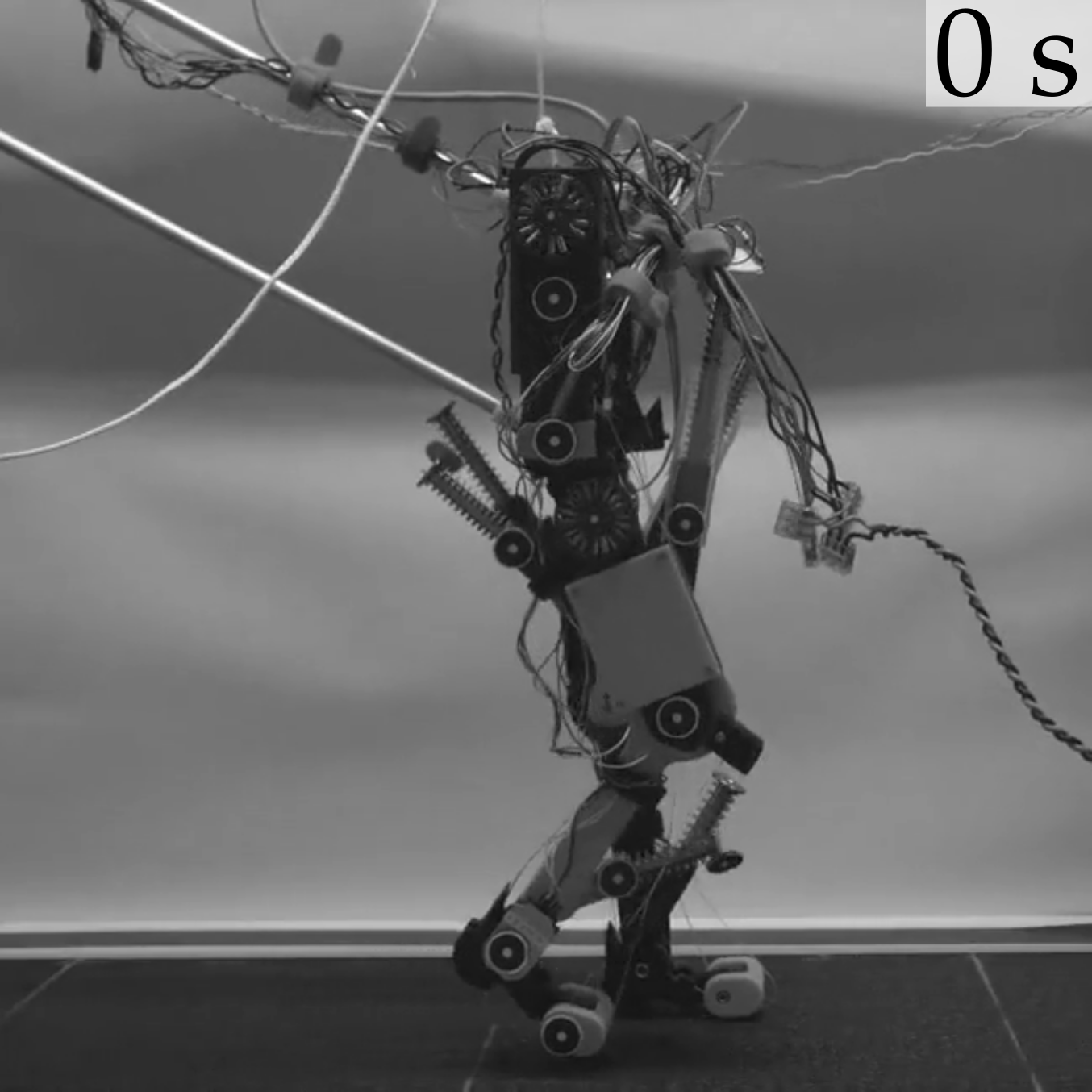} \end{subfigure}
\begin{subfigure}[b]{.30\textwidth} \includegraphics[width=.9\textwidth]{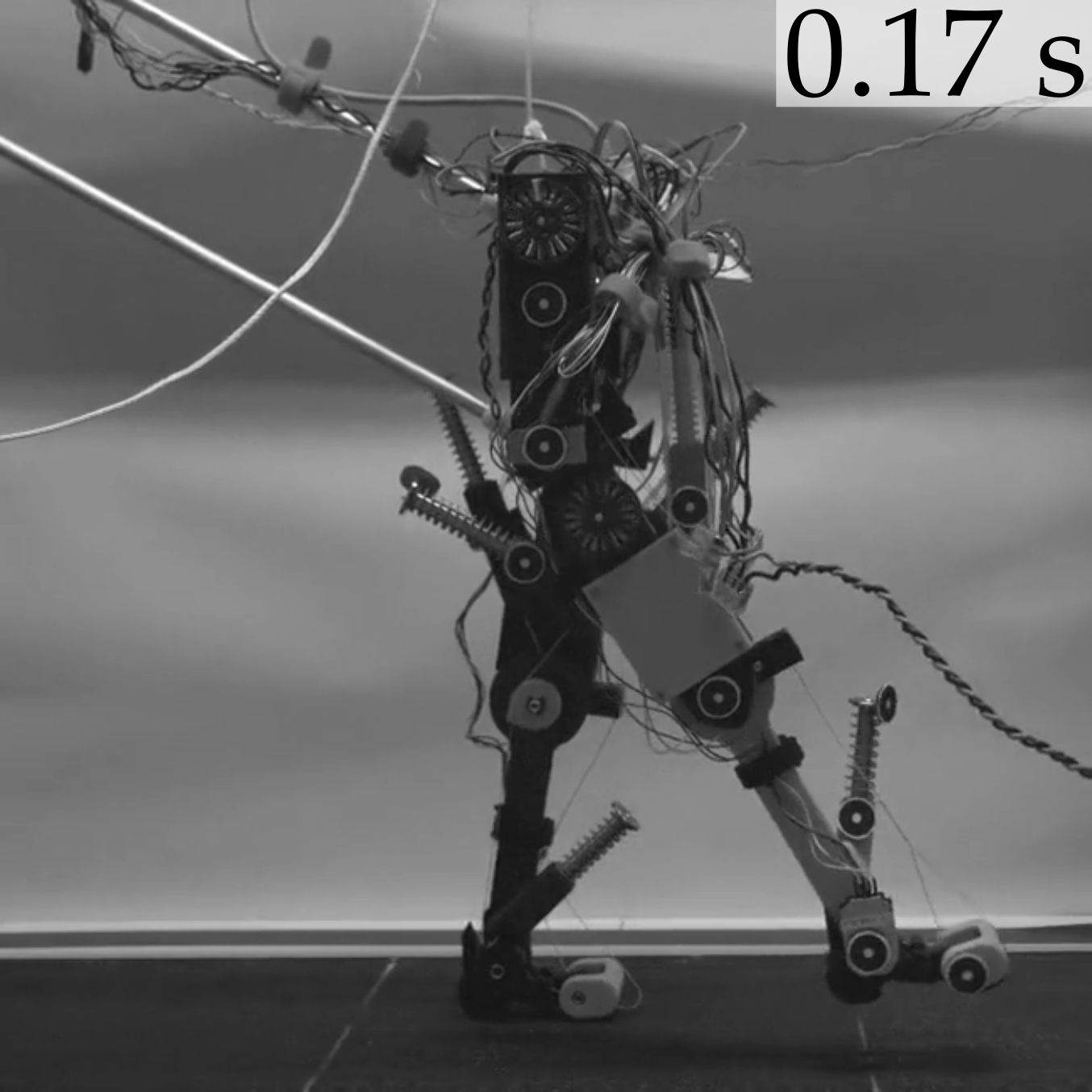} \end{subfigure}
\begin{subfigure}[b]{.30\textwidth} \includegraphics[width=.9\textwidth]{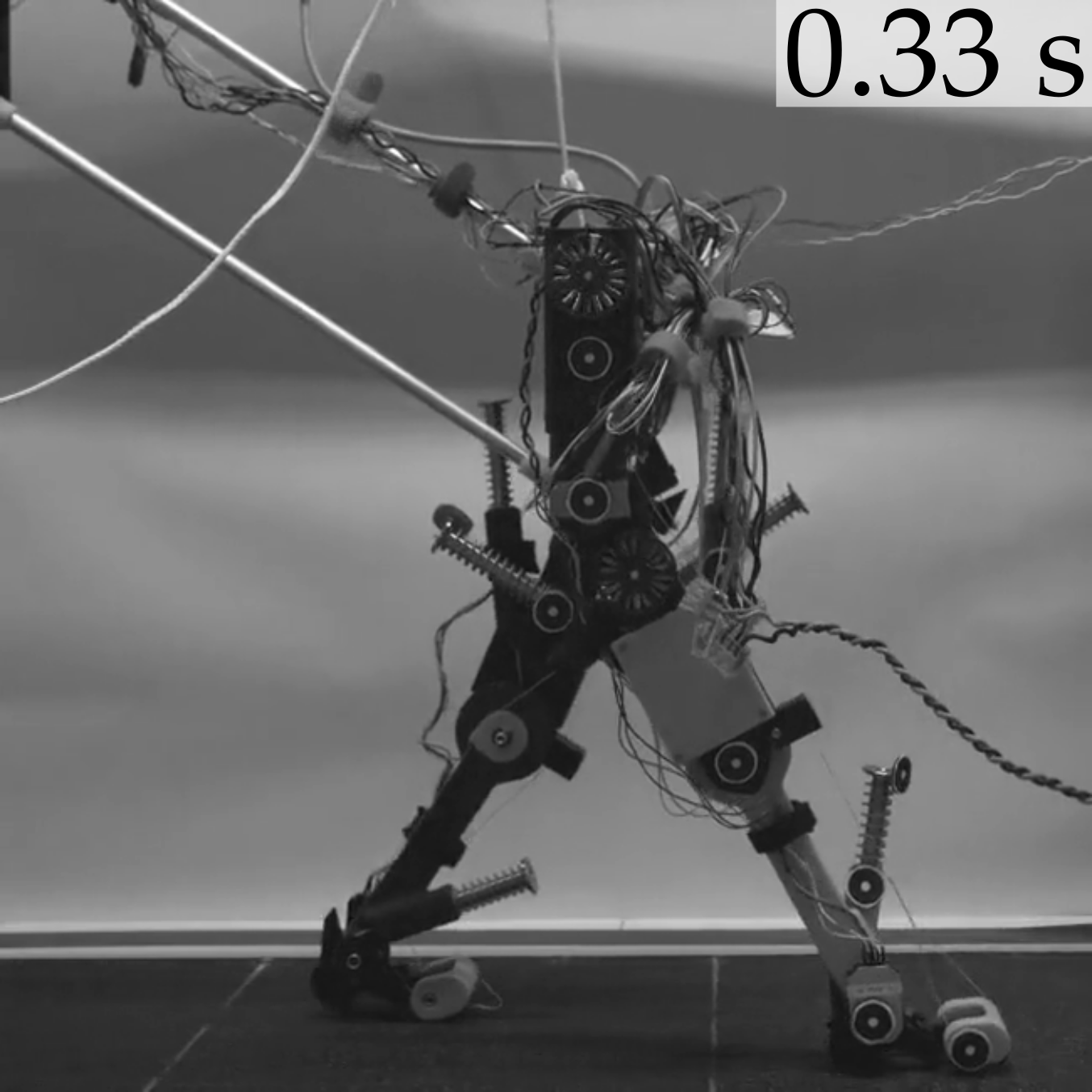} \end{subfigure}\\
\begin{subfigure}[b]{.30\textwidth} \includegraphics[width=.9\textwidth]{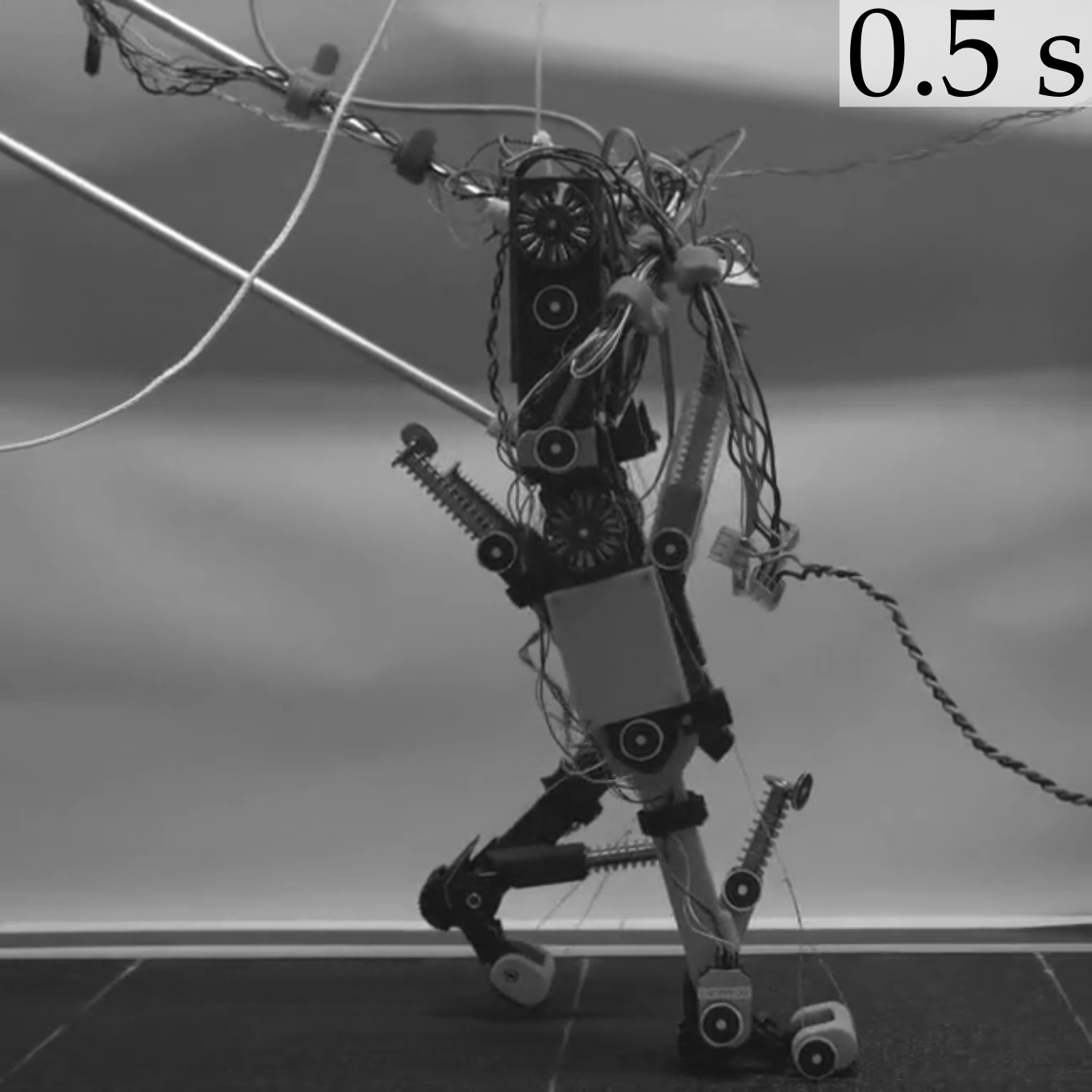} \end{subfigure}
\begin{subfigure}[b]{.30\textwidth} \includegraphics[width=.9\textwidth]{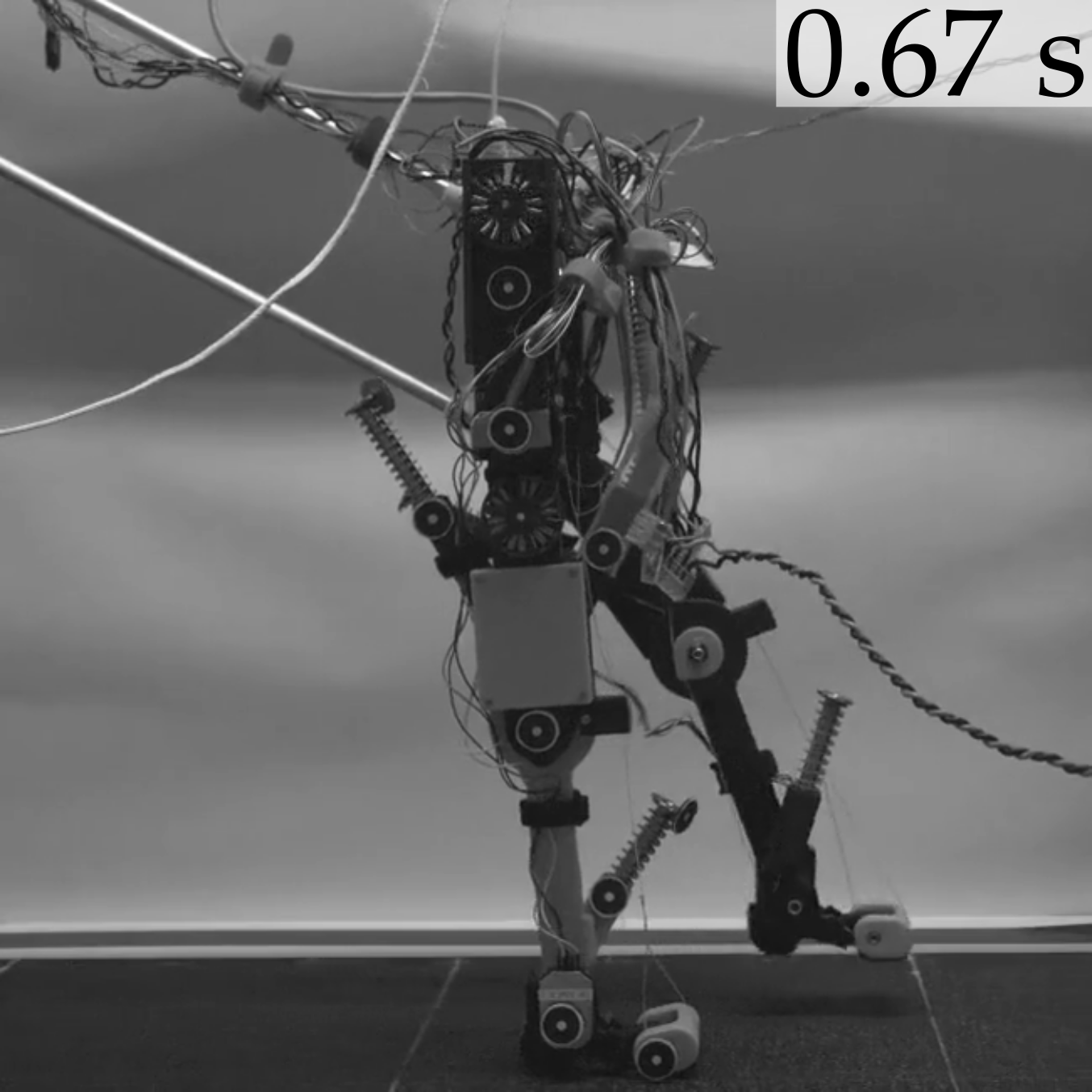} \end{subfigure}
\begin{subfigure}[b]{.30\textwidth} \includegraphics[width=.9\textwidth]{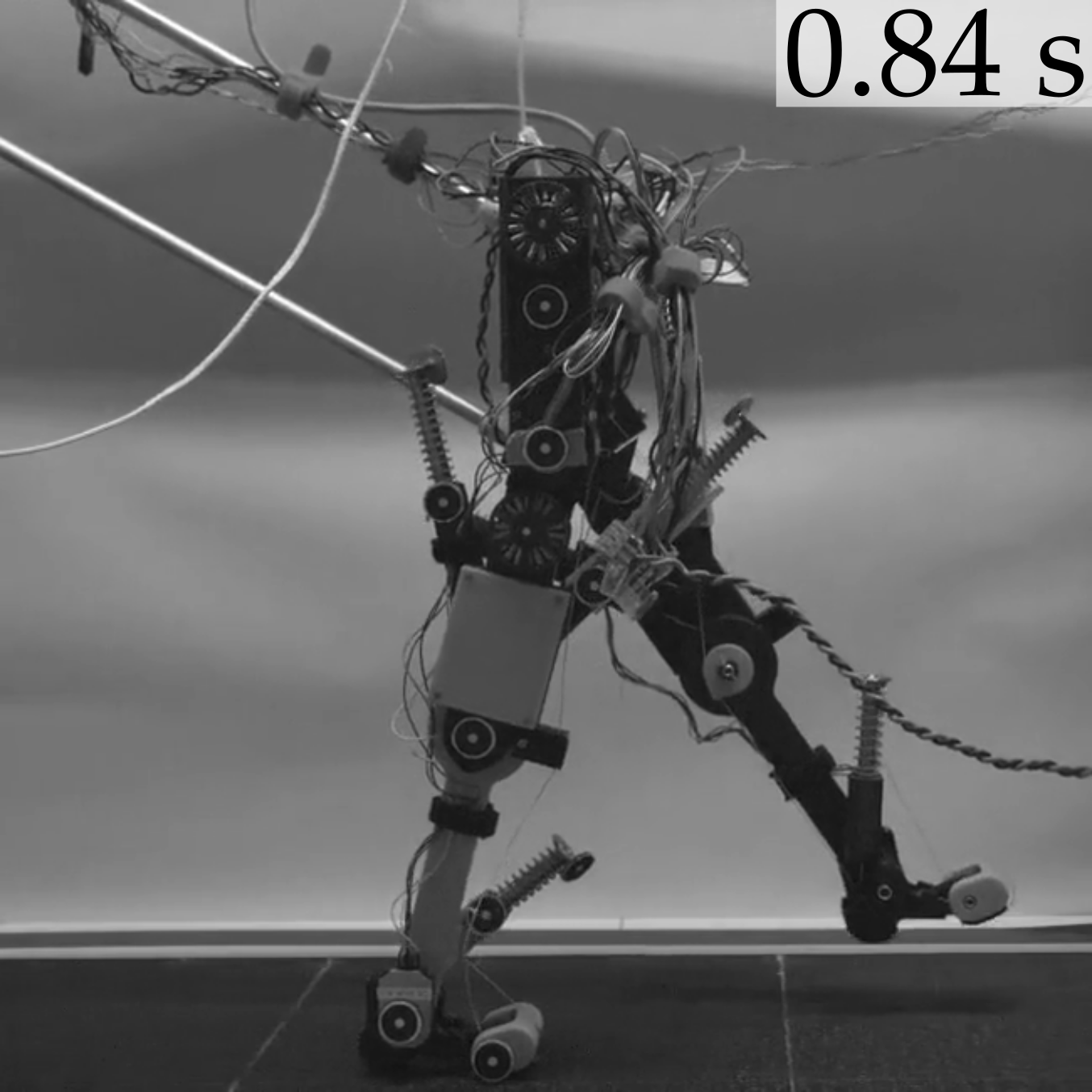} \end{subfigure}
\caption{Snapshots of $GAS+SOL$ configuration during one gait cycle. In-air swing leg retraction is visible in snapshot 2 and 6. One entire gait cycle lasts \SI{1.0}{s}.}
\label{fig:snapshots}
\end{figure*}

The robot walked on a treadmill with a gait cycle frequency of \SI{1.0}{Hz} in all three configurations. A gait with a \SI{25}{\degree} hip joint amplitude and \SI{0.35}{m} leg length leads to a locomotion speed of \SI{0.6}{m/s} (\SI{1.71}{leg-length/s}), at \SI{1.0}{Hz} locomotion frequency. This is \SI{50}{\%} of the preferred transition speed between walking and running. We calculated the robot's preferred transition speed as two thirds of the theoretically maximum walking speed at Froude number $Fr=$\,\num{1} \cite{Rotstein2005,Diedrich1995}. The robot's trunk was constrained to translations in the sagittal plane, and all trunk rotations were locked with a four-bar mechanism in all experiments (\Cref{fig:model,fig:snapshots}).

\subsection{Data acquisition and analysis}
% about data recording (sensors) and processing
% analysis methods
Our robot was instrumented for kinematics and power measurement.
We used rotary encoders at the hip and knee joints (AEDT-9810, Broadcom, 5000 CPR), at the four-bar mechanism, at the ankle joint (AMT-102, CUI, 4096 CPR), and on the treadmill (AEAT8800, Broadcom, 4096 CPR).
% The hip and knee angles are measured with rotary encoders.
% The four-bar mechanism and ankle angles are measured with rotary encoders (AMT-102, CUI, 4096 CPR).
% The robot forward motion was measured with the linear potentiometer on the four-bar slider and the treadmill encoder
% The treadmill speed was measured with a rotary encoder (AEAT8800, Broadcom, 4096 CPR).
The power consumption for hip and knee motor was measured by current sensors (ACHS-7121, Broadcom).
All sensors were sampled around \SI{600}{Hz} by a single board computer (4B, \text{Raspberry Pi Foundation}). Since sampling intervals were not constant, sensor data were interpolated to \SI{1000}{Hz} before analysis.

For each robot configuration (\Cref{tab:robot_para}), we commanded the robot to walk on the treadmill for \SI{120}{\second}.
The collected data was trimmed to the \SI{100}{\second} (100 cycles) where the robot's gait was in steady state.
% The current reading and ankle angle reading were filtered with moving average filter (window size = \SI{10}{ms}) while the rest data were kept in raw reading.
We averaged the gait cycles starting from touch-down without any filtering. Touch-down was defined as the moment where the ankle joint starts dorsiflexing after swing.

The ankle joint power was calculated using the SOL and GAS spring stiffness values, and the ankle and knee angles: 
\begin{equation}
\mathcal{P}_{A} = 
\begin{cases}
\begin{array}{l}
\omega_A \cdot (k_{SOL} \cdot r^2_{SOL} \cdot \alpha_A +... \\  k_{GAS} \cdot r^2_{GAS} \cdot (\alpha_A - \alpha_K)),
\vspace{0.5cm}
\end{array} & \text{if } \alpha_A \geq \alpha_K\\
\begin{array}{l}
\omega_A \cdot k_{SOL} \cdot r^2_{SOL} \cdot \alpha_A,
\end{array} & \text{else}
\end{cases}
\end{equation}
where $\mathcal{P}_{A}$ is the ankle joint power, $\omega_A$ is the angular velocity of the ankle in \SI{}{rad/s}, $\alpha_A$ and $\alpha_K$ are angle values of the ankle and knee joints in \SI{}{rad}, $k_{SOL}$ and $k_{GAS}$ are spring stiffnesses of SOL and GAS respectively in \SI{}{N/m}, $r_{SOL}$ and $r_{GAS}$ are pulley radii of SOL and GAS respectively in \SI{}{m}.
%------------------------------------------------------------------------------
%RESULTS:
%------------------------------------------------------------------------------
\section{RESULTS}

\begin{figure}[!h]
\centering
\includegraphics[width=\linewidth]{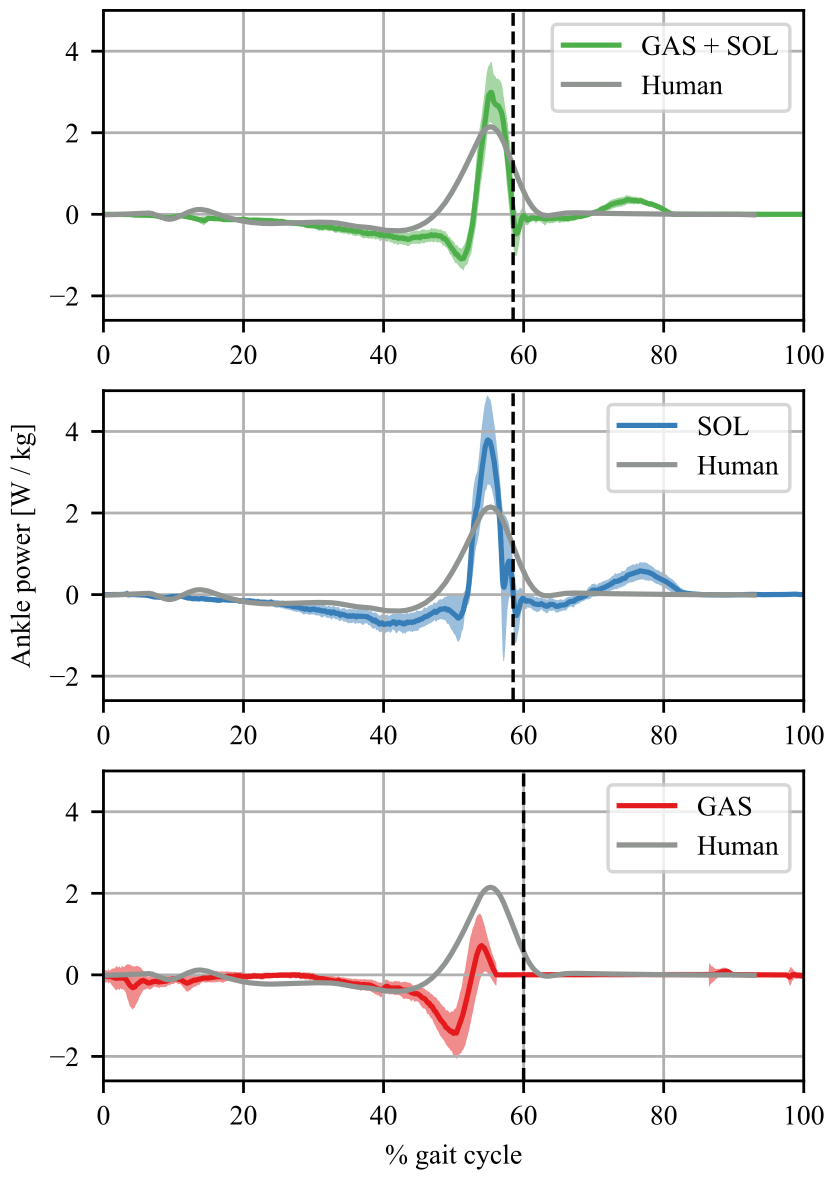}
\caption{Ankle joint power curves of three experimental configurations (in color) and human reference data (in gray, from \cite{Lipfert2014}). Shading shows the standard deviation of the curves. The vertical dashed lines indicate toe-off.}
\label{fig:power}
\vspace{-0.5cm}
\end{figure}

\begin{figure}[!h]
\centering
\includegraphics[width=\linewidth]{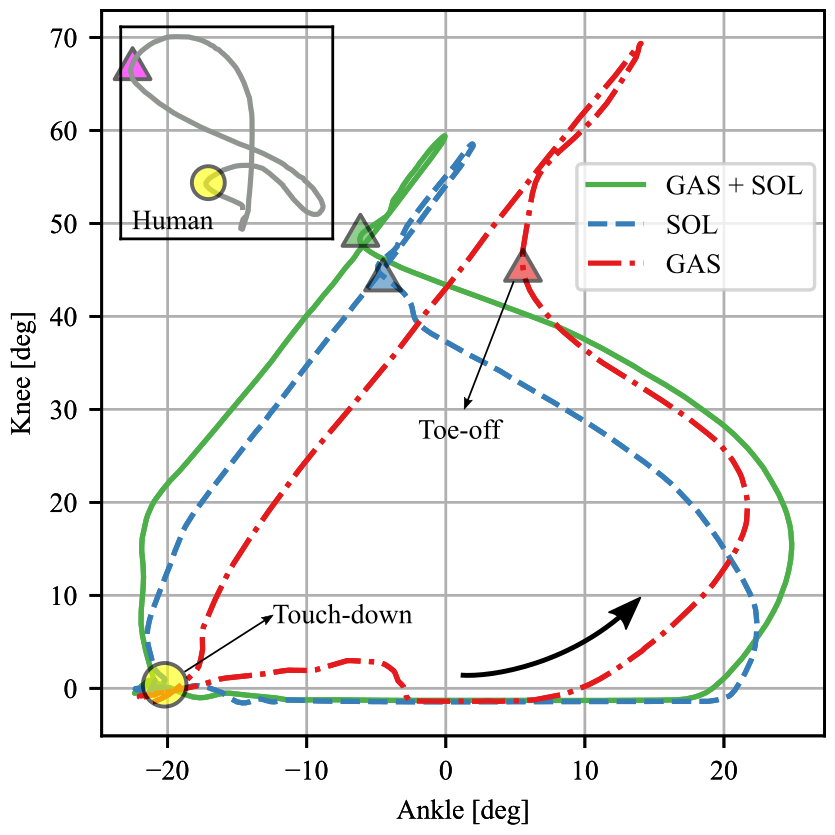}
\caption{Knee vs.\,ankle phase plot of the three experimental configurations(in color) and human reference data (in gray, from \cite{Lipfert2014}). The phase plot reads counterclockwise, starting from the touchdown (circle) to the toe-off (triangle) at the end of stance. Substantial differences are visible between the three configurations.}
\label{fig:ankle_knee}
\vspace{-0.5cm}
\end{figure}

% figure snapshots
\begin{table}[b]
\caption{Robot energetics. Ankle power amplification (ratio of positive and negative power peaks), walking speed, total positive-power CoT, and net positive-power CoT values for the three configurations.}
\label{tab:cot}
\begin{center}
\begin{tabular}{L{3.5cm}C{1.5cm}C{0.9cm}C{0.9cm}}
 						& GAS + SOL    & SOL	  &  GAS 	\\
\hline
Ankle power amplification [-] & 2.73 & 5.16 & 0.50 \\
Walking speed [m/s]    	& 0.55         & 0.57     & 0.35 	\\
Total positive-power COT [-]   		& 1.50         & 1.47     & 2.93    \\
Net positive-power COT [-]   	& 0.74         & 0.68     & 1.63 	\\
\end{tabular}
\end{center}
\end{table}
Joint angles show no large differences between the $GAS+SOL$ and $SOL$ configurations. For the $GAS$ configuration the ankle joint dorsiflexes sooner than in the other configurations and the ankle plantarflexes less during push-off (\Cref{fig:angle}). The main differences between the robot and human data are visible in ankle joint angles.

Similarly, the ankle joint power curve does not change between the $GAS+SOL$ and $SOL$ configurations, while for the $GAS$ configuration the negative power peak was larger and the positive power peak was lower than in the other two configurations (\Cref{fig:power}). Power amplification values of the ankle joint power (ratio of positive and negative power peaks) (\Cref{fig:power}) were \num{2.7}, \num{5.2} and \num{0.5} in the $GAS+SOL$, $SOL$ and $GAS$ configuration, respectively (\Cref{tab:cot}). Positive peaks occur at \SI{54}{}-\SI{55}{\%} of the gait cycle in all three robot configurations. Negative peaks occured at \SI{50}{}-\SI{51}{\%} in $GAS+SOL$ and $GAS$ configurations, and at \SI{40}{\%} in the $SOL$ configuration. Compared to the human data, the positive power peak in the $SOL$ configuration was almost twice as large as in the humans but in the $GAS+SOL$ configuration the positive peak height was close to identical to the human. A large negative peak appears in stance before push-off when $GAS$ was in the configurations, while this peak does not show in human data (\Cref{fig:power}).

The segment motion coordination (\Cref{fig:ankle_knee}) shows differences between all three configurations. In the $SOL$ configuration the ankle starts plantarflexing in late stance when the knee starts flexing while in the other two configurations the ankle dorsiflexes \SI{5}{\degree} and \SI{10}{\degree} more, respectively, after the knee starts flexing. The ankle push-off starts at \SI{5}{\degree}, \SI{15}{\degree} and \SI{20}{\degree} knee flexion in the $SOL$, $GAS+SOL$ and $GAS$ configurations, respectively. In the $GAS$ configuration the knee starts flexing in late stance at \SI{8}{\degree} ankle \df{} while the knee flexes only at $\sim$\SI{20}{\degree} ankle \df{} by the other two configurations (\Cref{fig:ankle_knee}).

We recorded the robot's total power consumption of \SI{18}{W}, \SI{17}{W}, and \SI{20}{W}, and walking speed of \SI{0.55}{m/s}, \SI{0.57}{m/s} and \SI{0.35}{m/s}, for configurations $GAS+SOL$, $SOL$, and $GAS$, respectively. The power values include the robot's motor drivers and its robot-mounted master control board, which draw in sum $\sim$\SI{9}{W} standby power.
We calculate the robot's total cost of transport (CoT) as $CoT=P/(mgv)$, with $P$ as the average total positive power (negative power values removed) measured including all actuators and motor drivers \cite{Seok.2013,Sprowitz.2013}, $m$ as the robot's mass, $g$ as the gravitational acceleration (\SI{9.81}{m/s^2}), and $v$ as the average horizontal robot speed over \num{100} gait cycles. Therefore, the humanoid robot's total, positive-power CoT values were \SI{1.5}{}, \SI{1.5}{} and \SI{2.9}{} for configurations $GAS+SOL$, $SOL$ and $GAS$, respectively. We find no major difference between configuration $GAS+SOL$ and $SOL$, but a \SI{93}{\%} larger CoT value for the $GAS$ configuration, compared to both other configurations (\Cref{tab:cot}).
%------------------------------------------------------------------------------
%DISCUSSION:
%------------------------------------------------------------------------------
 \section{DISCUSSION}
% summary of main results:
We show the differences in the function of SOL and GAS MTUs, represented by linear springs, in the ankle push-off during walking gait using three experimental configurations on an anthropomorphic bipedal robot. Ankle power amplification is visible in two of the experimental configurations (\Cref{fig:power}). We see differences between the configurations not only for the ankle power curves but also in the motion coordination of the ankle and knee joints (\Cref{fig:ankle_knee}). The SOL spring-tendon provides ankle power amplification while the GAS spring-tendon changes the movement coordination between ankle and knee joints during push-off.

% walking speed and CoT comparison between the 3 cases - advantage of SOL
Walking speed and CoT did not change substantially between the $GAS+SOL$ and $SOL$ configurations, the latter performed slightly better. Even though the power amplification was higher in the $SOL$ configuration and the ankle-knee movement coordination was different, the energetic consumption did not change substantially. Higher power amplification means that the elastic energy stored in the springs was released faster than it got stored. The loading of the springs was slower than the unloading which makes the leg of the robot comparable to a catapult in the $GAS+SOL$ and $SOL$ configurations. In the $GAS$ configuration, the ratio of positive and negative power peaks was smaller than one, i.e., there was no power amplification (\Cref{fig:power}, \Cref{tab:cot}). In accordance, speed decreased and CoT increased in the $GAS$ configuration compared to the other two configurations. Our robotic results support the understanding that the $SOL$ muscle is the major elastic element providing energy for the ankle push-off during walking as shown in human subject studies \cite{Fukunaga.2001,Ishikawa.2005,Lichtwark.2006,Lichtwark.2007} and studies that emulate $SOL$ function by springs with clutches in ankle-foot orthoses \cite{Collins.2015,Sawicki.2009}.

% CoT comparison to the natural runner 
The CoT values of the $GAS+SOL$, $SOL$ and $GAS$ configurations were \SI{55}{\%}, \SI{50}{\%} and \SI{119}{\%} of the approximated CoT of a natural walker with the same weight as these configurations (calculation method in \cite{birdbot_2022}). Many factors could influence the CoT, a freely walking robot on natural terrain and with trunk control would probably consume more power.

% CPG changes and reasons for these changes between the 3 cases
While trying to reach a stable gait pattern with the robot in the different configurations, the CPG parameters had to be minimally re-tuned and the ankle-knee joint angle coordination changed. Gait pattern changes were inevitable as the main difference between the SOL and GAS spring-tendons is that one spans only the ankle joint while the other spans both the ankle and knee joints. 

To reach a stable walking pattern with the robot in $SOL$ configuration, only the hip offset had to be decreased by \SI{2.5}{\degree} compared to the $GAS+SOL$ configuration. Before this change in the $SOL$ configuration, the stride length of the robot became too large, the contralateral leg hit the ground as it was starting to swing. This effect of decreasing GAS stiffness on the stride length was also shown in a simulation study \cite{Falisse2022}. 

In the $GAS$ configuration, without a $SOL$ spring, the knee buckled too early at the end of stance because the motors could not keep the knee straight for larger ankle \df{} angles. To counter large \df{} angles, the duty factor of the knee was decreased by \SI{2.5}{\%} of the gait cycle, in order to start knee flexion at the end of stance earlier than the moment of maximum hip extension. Additionally, we decreased hip amplitude and offset both by \SI{2.5}{\degree} compared to the $GAS+SOL$ configuration. The changes of the CPG parameters are apparent in the ankle-knee joint angle coordination (\Cref{fig:ankle_knee}) as well. The knee starts flexing at \SI{8}{\degree} ankle dorsiflexion angle in the $GAS$ configuration instead of \SI{18}{\degree} dorsiflexion in the $GAS+SOL$ configuration.
% ankle-knee coordination changes - advantage of GAS
The biarticular $GAS$ spring-tendon apparently changes the ankle-knee joint coordination which is an important aspect of human ankle push-off during walking \cite{Schumacher.2020} because it affects the energy transfer between ankle and knee joints.

% Limitations + future direction
The robot's toe clearance in swing was a limiting factor, which made larger hip amplitudes necessary. Another limiting factor was that hip and knee joints were CPG position-controlled during the whole gait cycle. Consequently, the natural springy knee flexion of the human was missing in stance, and the robot's ankle joint angles show clear differences in all three configurations compared to humans'. That is, the active control of the knee and hip joints affects the natural ankle, knee and hip dynamics induced by the energy release during ankle push-off. In the future, switching knee and hip actuators to low gain control in selected parts of the gait cycle could potentially lead to more natural swing leg dynamics.

To reach our goal to separate the roles of SOL and GAS spring-tendons, our quantitative analysis was limited to three set of chosen spring stiffness values and rest lengths. In the future, multiple SOL-GAS spring stiffness ratios and the influence of spring-tendon rest-length should be analysed for a more comprehensive picture on the role of SOL and GAS spring-tendons.

% Interpret your results. Is the performance acceptable? Go back to initial questions, compare to other methods reported in literature. How do you explain unexpected results? Which shortcomings or limitations are there in your methods and results? Where is room for improvement? Could combination with methods from literature further improve the work? Mention plans, needs, and recommendations for future research. %This may also be a separate subsection ``Future Work''.
% Literature that may be interesting to compare against
% \cite{Collins.2015}
% \cite{Sawicki.2009}
%------------------------------------------------------------------------------
%CONCLUSION:
%------------------------------------------------------------------------------
\section{CONCLUSION}
In conclusion, differences can be found between the robot configurations using both SOL and GAS springs, only SOL spring-tendon, and only GAS spring-tendon in the ankle power curves, the movement coordination of the ankle and knee joints, cost of transport, and walking speed. Robotic implementation of the SOL and GAS muscle-tendon units made it possible to analyse their separate roles in push-off in a simplified setup. The SOL spring-tendon provides ankle power amplification seen in the ankle power curve, likely allowing lower cost of transport and higher walking speed, while the GAS spring-tendon changes the movement coordination between ankle and knee joints during push-off. In the future, more natural swing leg dynamics should be investigated by switching between low and high gain control of the actuators during a gait cycle. We are further interested in actuating the currently passive ankle mechanism by mechanically efficient, series elastic actuation \cite{bolignari_diaphragm} to adjust push-off timing.

% This is not a summary. Instead, state the take-home message.
% A conclusion section is not required. Although a conclusion may review the main points of the paper, do not replicate the abstract as the conclusion. A conclusion might elaborate on the importance of the work or suggest applications and extensions. 
%\addtolength{\textheight}{-12cm}   % This command serves to balance the column lengths
                                  % on the last page of the document manually. It shortens
                                  % the textheight of the last page by a suitable amount.
                                  % This command does not take effect until the next page
                                  % so it should come on the page before the last. Make
                                  % sure that you do not shorten the textheight too much.
%------------------------------------------------------------------------------
%APPENDIX:
%------------------------------------------------------------------------------
\section*{APPENDIX}
Video footage of the anthropomorphic bipedal robot's locomotion on the treadmill in all three configurations is available on YouTube \url{https://youtu.be/T79pKLQ47XU} and through the submitted supplementary video file.

Experimental data and visualization code are available at \url{https://doi.org/10.17617/3.BQ2PZ9}. CAD design files and robot control code are available for non-commercial use at the same link.

%------------------------------------------------------------------------------
%ACKNOWLEDGMENTS:
%------------------------------------------------------------------------------
\section*{ACKNOWLEDGMENTS}
This work was funded by the Deutsche Forschungsgemeinschaft (DFG, German Research Foundation) 449427815, 449912641. The authors thank the International Max Planck Research School for Intelligent Systems (IMPRS-IS) for supporting Bernadett Kiss, Emre Cemal Gonen, and An Mo. An Mo was also supported by the China Scholarship Council.

The authors would like to thank Alborz Aghamaleki Sarvestani and Felix Ruppert for their helpful advice during the design process; Felix Grimminger and his team for the help with the actuator modules; the Robotics ZWE at MPI-IS for the 3D-printed parts, and Guido Nafz from the MPI-IS Precision Mechanics Workshop for the precision parts.
% Thank everyone who contributed with smaller pieces of advice or non-scientifically (e.g.\ technical support).
% Example: The authors would like to thank [...]
% \vspace*{20cm}
\bibliography{literature}

\begin{thebibliography}{10}
\providecommand{\url}[1]{#1}
\csname url@rmstyle\endcsname
\providecommand{\newblock}{\relax}
\providecommand{\bibinfo}[2]{#2}
\providecommand\BIBentrySTDinterwordspacing{\spaceskip=0pt\relax}
\providecommand\BIBentryALTinterwordstretchfactor{4}
\providecommand\BIBentryALTinterwordspacing{\spaceskip=\fontdimen2\font plus
\BIBentryALTinterwordstretchfactor\fontdimen3\font minus
  \fontdimen4\font\relax}
\providecommand\BIBforeignlanguage[2]{{%
\expandafter\ifx\csname l@#1\endcsname\relax
\typeout{** WARNING: IEEEtran.bst: No hyphenation pattern has been}%
\typeout{** loaded for the language `#1'. Using the pattern for}%
\typeout{** the default language instead.}%
\else
\language=\csname l@#1\endcsname
\fi
#2}}

\bibitem{Lipfert2014}
S.~Lipfert, M.~G{\"{u}}nther, D.~{Renjewski}, and A.~Seyfarth, ``{Impulsive
  ankle push-off powers leg swing in human walking},'' \emph{The Journal of
  experimental biology}, vol. 217, pp. 1218--1228, 2014.

\bibitem{Hof1983}
A.~L. Hof, B.~A. Geelen, and J.~{Van den Berg}, ``{Calf muscle moment, work and
  efficiency in level walking; Role of series elasticity},'' \emph{Journal of
  Biomechanics}, vol.~16, no.~7, pp. 523--537, 1983.

\bibitem{Fukunaga.2001}
T.~Fukunaga, K.~Kubo, Y.~Kawakami, S.~Fukashiro, H.~Kanehisa, and C.~N.
  Maganaris, ``In vivo behaviour of human muscle tendon during walking,''
  \emph{Proceedings. Biological Sciences}, vol. 268, no. 1464, pp. 229--233,
  2001.

\bibitem{Ishikawa.2005}
M.~Ishikawa, P.~V. Komi, M.~J. Grey, V.~Lepola, and G.-P. Bruggemann,
  ``Muscle-tendon interaction and elastic energy usage in human walking,''
  \emph{Journal of Applied Physiology (Bethesda, Md. : 1985)}, vol.~99, no.~2,
  pp. 603--608, 2005.

\bibitem{Alexander.1991}
R.~M. Alexander, ``Energy-saving mechanisms in walking and running,'' \emph{The
  Journal of Experimental Biology}, vol. 160, no.~1, pp. 55--69, 1991.

\bibitem{Lichtwark.2006}
G.~A. Lichtwark and A.~M. Wilson, ``Interactions between the human
  gastrocnemius muscle and the achilles tendon during incline, level and
  decline locomotion,'' \emph{The Journal of Experimental Biology}, vol. 209,
  no. Pt 21, pp. 4379--4388, 2006.

\bibitem{Lichtwark.2007}
G.~A. Lichtwark, K.~Bougoulias, and A.~M. Wilson, ``Muscle fascicle and series
  elastic element length changes along the length of the human gastrocnemius
  during walking and running,'' \emph{Journal of Biomechanics}, vol.~40, no.~1,
  pp. 157--164, 2007.

\bibitem{Cronin.2013c}
N.~J. Cronin, J.~Avela, T.~Finni, and J.~Peltonen, ``Differences in contractile
  behaviour between the soleus and medial gastrocnemius muscles during human
  walking,'' \emph{Journal of Experimental Biology}, vol. 216, no. Pt 5, pp.
  909--914, 2013.

\bibitem{Cronin.2013}
N.~J. Cronin, B.~I. Prilutsky, G.~A. Lichtwark, and H.~Maas, ``Does ankle joint
  power reflect type of muscle action of soleus and gastrocnemius during
  walking in cats and humans?'' \emph{Journal of Biomechanics}, vol.~46, no.~7,
  pp. 1383--1386, 2013.

\bibitem{Sawicki.2009}
G.~S. Sawicki, C.~L. Lewis, and D.~P. Ferris, ``It pays to have a spring in
  your step,'' \emph{Exercise and Sport Sciences Reviews}, vol.~37, no.~3, pp.
  130--138, 2009.

\bibitem{Collins.2015}
S.~H. Collins, M.~B. Wiggin, and G.~S. Sawicki, ``Reducing the energy cost of
  human walking using an unpowered exoskeleton,'' \emph{Nature}, vol. 522, no.
  7555, pp. 212--215, 2015.

\bibitem{sharbafi_new_2016}
M.~A. Sharbafi, C.~Rode, S.~Kurowski, D.~Scholz, R.~Möckel, {Katayon Radkhah},
  G.~Zhao, A.~M. Rashty, O.~v. Stryk, and A.~Seyfarth, ``A new biarticular
  actuator design facilitates control of leg function in {BioBiped3},''
  \emph{Bioinspiration \& Biomimetics}, vol.~11, no.~4, p. 046003, 2016.

\bibitem{nejadfard_design_2019}
A.~Nejadfard, S.~Schütz, K.~Mianowski, P.~Vonwirth, and K.~Berns, ``Design of
  the musculoskeletal leg based on the physiology of mono-articular and
  biarticular muscles in the human leg,'' \emph{Bioinspiration \& biomimetics},
  vol.~14, no.~6, p. 066002, 2019.

\bibitem{eslamy_adding_2014}
M.~Eslamy, M.~Grimmer, and A.~Seyfarth, ``Adding passive biarticular spring to
  active mono-articular foot prosthesis: {Effects} on power and energy
  requirement,'' in \emph{2014 14th {IEEE}-{RAS} {International} {Conference}
  on {Humanoid} {Robots} ({Humanoids})}, Nov. 2014, pp. 677--684.

\bibitem{birdbot_2022}
A.~Badri-Spröwitz, A.~Aghamaleki~Sarvestani, M.~Sitti, and M.~A. Daley,
  ``{BirdBot} achieves energy-efficient gait with minimal control using
  avian-inspired leg clutching,'' \emph{Science Robotics}, vol.~7, no.~64, p.
  eabg4055, Mar. 2022.

\bibitem{Plagenhoef1983}
S.~Plagenhoef, F.~{Gaynor Evans}, and T.~Abdelnour, ``{Anatomical Data for
  Analyzing Human Motion},'' \emph{Research Quarterly for Exercise and Sport},
  vol.~54, no.~2, pp. 169--178, 1983.

\bibitem{Gharooni2007}
S.~Gharooni, S.~Sareh, and M.~O. Tokhi, ``{Development of FES-rowing machine
  with quadriceps stimulation and energy storing device},'' no. November, pp.
  1--3, 2007.

\bibitem{ijspeert_central_2008}
A.~J. Ijspeert, ``Central pattern generators for locomotion control in animals
  and robots: {A} review,'' \emph{Neural Networks}, vol.~21, no.~4, pp.
  642--653, May 2008.

\bibitem{Grimminger2019}
F.~Grimminger, A.~Meduri, M.~Khadiv, J.~Viereck, M.~Wuthrich, M.~Naveau,
  V.~Berenz, S.~Heim, F.~Widmaier, T.~Flayols, J.~Fiene, A.~Badri-Sprowitz,
  L.~Righetti, M.~W{\"{u}}thrich, M.~Naveau, V.~Berenz, S.~Heim, F.~Widmaier,
  J.~Fiene, A.~Badri-Spr{\"{o}}witz, and L.~Righetti, ``An open
  force-controlled modular robot architecture for legged locomotion research,''
  \emph{IEEE Robotics and Automation Letters}, vol.~5, no.~2, pp. 3650--3657,
  April 2020.

\bibitem{Geyer.2010}
H.~Geyer and H.~Herr, ``A muscle-reflex model that encodes principles of legged
  mechanics produces human walking dynamics and muscle activities,'' \emph{IEEE
  transactions on neural systems and rehabilitation engineering : a publication
  of the IEEE Engineering in Medicine and Biology Society}, vol.~18, no.~3, pp.
  263--273, 2010.

\bibitem{ashtiani_hybrid_2021}
M.~S. Ashtiani, A.~A. Sarvestani, and A.~T. Badri-Spröwitz, ``Hybrid parallel
  compliance allows robots to operate with sensorimotor delays and low control
  frequencies,'' \emph{Frontiers in Robotics and AI}, vol.~8, 2021.

\bibitem{Perry.2010}
J.~Perry, J.~M. Burnfield, and L.~M. Cabico, \emph{Gait Analysis: Normal and
  Pathological Function}.\hskip 1em plus 0.5em minus 0.4em\relax SLACK, 2010.

\bibitem{ijspeert2007swimming}
A.~J. Ijspeert, A.~Crespi, D.~Ryczko, and J.-M. Cabelguen, ``From swimming to
  walking with a salamander robot driven by a spinal cord model,''
  \emph{science}, vol. 315, no. 5817, pp. 1416--1420, 2007.

\bibitem{Delp1990}
S.~L. Delp, J.~P. Loan, M.~G. Hoy, F.~E. Zajac, E.~L. Topp, and J.~M. Rosen,
  ``{An Interactive Graphics-Based Model of the Lower Extremity to Study
  Orthopaedic Surgical Procedures},'' \emph{IEEE Transactions on Biomedical
  Engineering}, vol.~37, no.~8, pp. 757--767, 1990.

\bibitem{Rotstein2005}
A.~Rotstein, O.~Inbar, T.~Berginsky, and Y.~Meckel, ``{Preferred transition
  speed between walking and running: Effects of training status},''
  \emph{Medicine and Science in Sports and Exercise}, vol.~37, no.~11, pp.
  1864--1870, 2005.

\bibitem{Diedrich1995}
F.~J. Diedrich and W.~H. Warren, ``{Why Change Gaits? Dynamics of the Walk-Run
  Transition},'' \emph{Journal of Experimental Psychology: Human Perception and
  Performance}, vol.~21, no.~1, pp. 183--202, 1995.

\bibitem{Seok.2013}
S.~Seok, A.~Wang, M.~Y. Chuah, D.~Otten, J.~Lang, and S.~Kim, ``Design
  principles for highly efficient quadrupeds and implementation on the mit
  cheetah robot,'' in \emph{IEEE International Conference on Robotics and
  Automation (ICRA), 2013}.\hskip 1em plus 0.5em minus 0.4em\relax Piscataway,
  NJ: IEEE, 2013, pp. 3307--3312.

\bibitem{Sprowitz.2013}
A.~T. Badri-Spr{\"o}witz, A.~Tuleu, M.~Vespignani, M.~Ajallooeian, E.~Badri,
  and A.~J. Ijspeert, ``Towards dynamic trot gait locomotion: Design, control,
  and experiments with cheetah-cub, a compliant quadruped robot,'' \emph{The
  International Journal of Robotics Research}, vol.~32, no.~8, pp. 932--950,
  2013.

\bibitem{Falisse2022}
A.~Falisse, M.~Afschrift, and F.~{De Groote}, ``{Modeling toes contributes to
  realistic stance knee mechanics in three-dimensional predictive simulations
  of walking},'' \emph{PLOS ONE}, vol.~17, no.~1, p. e0256311, jan 2022.

\bibitem{Schumacher.2020}
C.~Schumacher, M.~Sharbafi, A.~Seyfarth, and C.~Rode, ``Biarticular muscles in
  light of template models, experiments and robotics: a review,'' \emph{Journal
  of The Royal Society Interface}, vol.~17, no. 163, p. 20180413, 2020.

\bibitem{bolignari_diaphragm}
M.~Bolignari, A.~Mo, M.~Fontana, and A.~Badri-Spr\"owitz, ``Diaphragm {Ankle}
  {Actuation} for {Efficient} {Series} {Elastic} {Legged} {Robot} {Hopping},''
  \emph{accepted for IROS2022 10.48550/arXiv.2203.01595}, p.~8.

\end{thebibliography}
\end{document}